\documentclass[sigconf, noacm]{acmart}
\AtBeginDocument{%
  }

\renewcommand\footnotetextcopyrightpermission[1]{}
\usepackage{multirow}
\usepackage{multicol}
\usepackage{colortbl}
\usepackage{tcolorbox}
\newcommand{\PredSty}[1]{\textnormal{\ttfamily\color{mygreen!90!black}#1}\unskip}
\definecolor{mygreen}{HTML}{3cb44b}
\usepackage{subcaption}

\usepackage{enumitem}

\begin{document}

\title[FEALLM: Advancing Facial Emotion Analysis in MLLMs with Emotional Synergy and Reasoning]{FEALLM: Advancing Facial Emotion Analysis in Multimodal Large Language Models with Emotional Synergy and Reasoning}


\author{Zhuozhao Hu$^{1}$*, Kaishen Yuan$^{2}$*, Xin Liu$^{3}$\textsuperscript{†}, Zitong Yu$^{4}$, Yuan Zong$^{5}$, \\ 
Jingang Shi$^{6}$, Huanjing Yue$^{1}$, Jingyu Yang$^{1}$  \\
\textsuperscript{1}Tianjin University  \textsuperscript{2}The Hong Kong University of Science and Technology (Guangzhou) \\
\textsuperscript{3}Lappeenranta-Lahti University of Technology LUT  \textsuperscript{4}Great Bay University \\
\textsuperscript{5}Southeast University  \textsuperscript{6}Xi'an Jiaotong University \\
}
\thanks{*Equal contributions.}
\thanks{\textsuperscript{†}Corresponding author.}
\email{huzhuozhao@tju.edu.cn, yuankaishen01@gmail.com, xin.liu@lut.fi}

\begin{abstract}
  Facial Emotion Analysis (FEA) plays a crucial role in visual affective computing, aiming to infer a person's emotional state based on facial data. Scientifically, facial expressions (FEs) result from the coordinated movement of facial muscles, which can be decomposed into specific action units (AUs) that provide detailed emotional insights. However, traditional methods often struggle with limited interpretability, constrained generalization and reasoning abilities. Recently, Multimodal Large Language Models (MLLMs) have shown exceptional performance in various visual tasks, while they still face significant challenges in FEA due to the lack of specialized datasets and their inability to capture the intricate relationships between FEs and AUs.
  To address these issues, we introduce a novel FEA Instruction Dataset that provides accurate and aligned FE and AU descriptions and establishes causal reasoning relationships between them, followed by constructing a new benchmark, FEABench.
  Moreover, we propose FEALLM, a novel MLLM architecture designed to capture more detailed facial information, enhancing its capability in FEA tasks.
  Our model demonstrates strong performance on FEABench and impressive generalization capability through zero-shot evaluation on various datasets, including RAF-DB, AffectNet, BP4D, and DISFA, showcasing its robustness and effectiveness in FEA tasks.
  The dataset and code will be available at https://github.com/953206211/FEALLM.
\end{abstract}





\maketitle

\begin{figure}[ht]
    \centering
    \includegraphics[width=\linewidth]{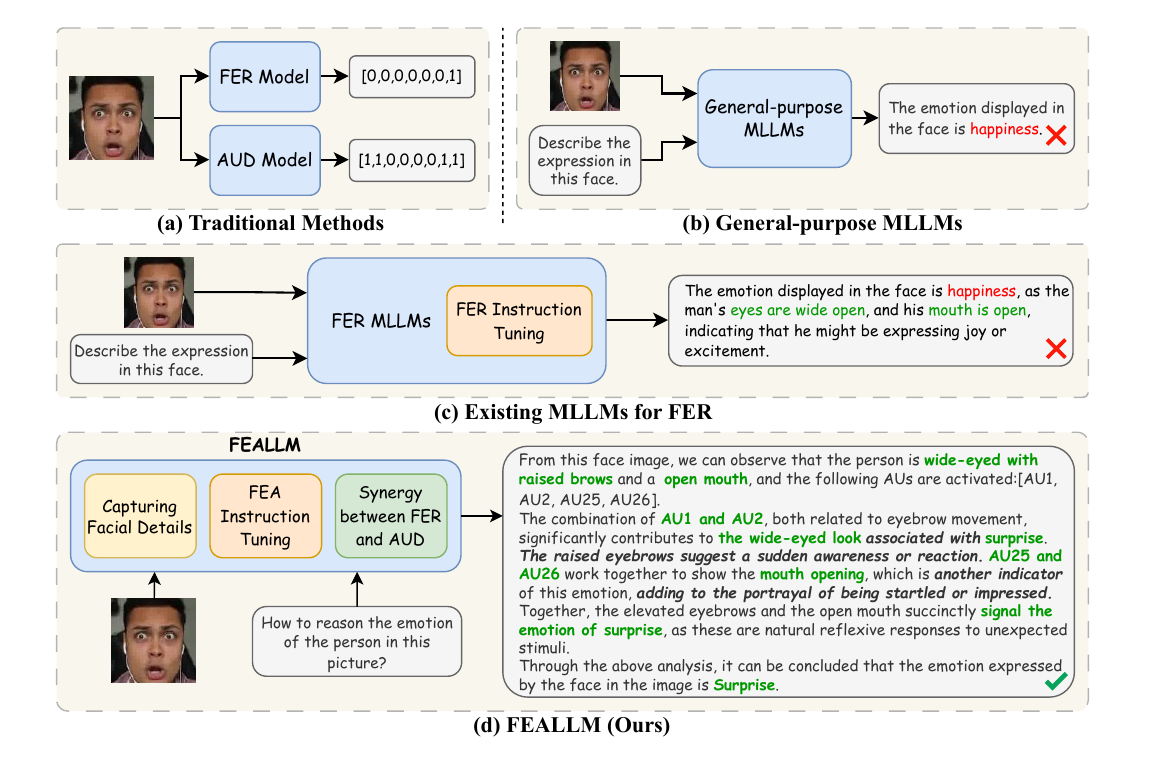}
    \vspace{-2.0em}
    \caption{Illustration of the innovativeness of the proposed FEALLM. (a) shows that traditional methods using binary labels result in a lack of interpretability and generalizability. (b) shows that existing general-purpose MLLMs typically exhibit poor facial emotion analysis capabilities. (c) shows that certain MLLMs designed for FER fail to link abstract emotions with facial movements, limiting the model's performance. (d) shows that our FEALLM is endowed with strong facial emotion perception and reasoning capabilities, owing to incorporating comprehensive FEA instruction tuning and capturing more detailed facial characteristics.}
    \label{fig1}
    \vspace{-1.0em}
\end{figure}

\section{Introduction}

Facial Emotion Analysis (FEA) aims to recognize and infer a person’s emotional state and inner activities based on facial data. Due to its broad range of applications, such as human-computer interaction~\cite{shi2020human,lu2024gpt} and mental health assessment~\cite{lee2021development,liu2021imigue}, it has long been a prominent research topic in the field of affective computing. Facial expression recognition (FER) and facial action unit detection (AUD) are two key tasks in FEA, used respectively to recognize facial expressions (FE), which are coarsely classified into seven categories~\cite{ekman2003darwin}, and facial action units (AU), which are finely defined by the Facial Action Coding System (FACS) based on the activation and movement of facial muscles~\cite{ekman1978facial}. FE and AU represent facial emotion at different levels, and various combinations of AUs can form distinct FEs, thus creating an intricate relationship between the two~\cite{martinez2017automatic}.

With the rapid development of deep learning, a large number of methods specifically designed for FER or AUD have emerged, typically treating them as multi-class or multi-label binary classification tasks~\cite{li2020deep,martinez2017automatic,AUTTT,MPSCL,ME-GraphAU,SRERL,MVT,RUL,Face2Exp,DAN}. While these traditional methods have achieved impressive recognition performance, they often rely on rigid discrete labels (e.g., binary labels) and are commonly designed to handle a single specific task. As a result, these methods often suffer from limited interpretability, as well as constrained generalization and reasoning abilities, as shown in Fig.~\ref{fig1} (a). \textit{These limitations reveal the urgent need for a more robust, comprehensive, and interpretable solution for FEA.}

Recently, Multimodal Large Language Models (MLLMs) ~\cite{bai2023qwen,chen2024expanding,li2025visual} have shown exceptional performance in various visual tasks such as visual question answering~\cite{hu2024bliva,sterner2024few,caffagni2024wiki} and image captioning~\cite{bianco2023improving,liu2023visual,li2023blip}, demonstrating strong reasoning and generalization capabilities, and providing promising solutions across various domains. Unfortunately, due to the lack of instruction datasets specifically designed for FEA in the training corpus, existing general-purpose MLLMs are still in the early stages of facial emotion perception, as shown in Fig.~\ref{fig1} (b). Some pioneering studies have attempted to introduce specialized corpora to enhance the FEA capabilities of MLLMs. However, most of these efforts primarily construct instruction datasets related to FER to guide the model, limiting MLLMs' ability to deeply understand abstract and fine-grained emotions~\cite{yang2024emollm,xing2024emo,cheng2024emotion}, as shown in Fig.~\ref{fig1} (c). An attempt has been made to introduce AUD-related instruction data; whereas, due to the misalignment of FE and AU annotations, FER and AUD are treated as separate tasks for training and evaluation, which prevents MLLMs from uncovering the intrinsic connections between emotional representations at different levels, thereby limiting their emotional reasoning capabilities~\cite{li2025facial}. Furthermore, most MLLMs focus on extracting global visual features when processing images, which may fail to capture the crucial local facial details necessary for detecting subtle facial emotions, limiting their performance in FEA. \textit{Therefore, an MLLM with stronger emotion perception capabilities and the ability to understand the intrinsic causal relationships between FE and AU remains an open research gap.}

Based on the observations above, we first construct a novel FEA instruction dataset. Specifically, this dataset not only includes more accurate and aligned basic FE and AU descriptions but also innovatively introduces reasoning instructions that bridge these two different granular levels of emotional representations. To the best of our knowledge, this is the first such effort, enabling MLLMs to understand how different facial muscle movements influence specific emotional states, thereby facilitating a deeper exploration of the intrinsic relationships between FE and AU and enhancing the model's reasoning abilities and interpretability in FEA. Furthermore, based on the constructed dataset, we introduce a new benchmark, FEABench. Unlike traditional emotion benchmarks that treat FER and AUD tasks as separate, FEABench is designed to simultaneously evaluate a model's performance in both FER and AUD tasks, enabling a more comprehensive assessment of MLLMs' capabilities in emotional perception and facial emotion analysis, and promoting the synergistic development of the model across these two tasks. Additionally, to address the common issue of insufficient detail capture, we specifically design a new MLLM for FEA, FEALLM, which enhances the attention to local facial details by extracting local features from facial images and combining them with low-level features from the visual encoder. Fig.~\ref{fig1} (d) demonstrates the advancement of our approach. Extensive experiments from multiple perspectives demonstrate the strong emotional perception, generalization, and reasoning abilities of the proposed FEALLM on FEABench, as well as on widely used FER datasets (RAF-DB~\cite{RAF-DB} and AffectNet~\cite{AffectNet}) and AUD datasets (BP4D~\cite{BP4D} and DISFA~\cite{DISFA}).

Our contributions can be summarized as follows:
\begin{itemize}
    \item We propose a novel \textbf{FEA Instruction Dataset}, which includes not only accurate and aligned basic FE and AU descriptions, but also innovative complex emotional reasoning instructions to facilitate the model's ability to perceive the intrinsic connections between FEs and AUs.
    \item We introduce \textbf{FEABench}, a new benchmark that simultaneously assesses model performance in both FER and AUD tasks, rather than treating them as separate, promoting synergistic development across these two tasks.
    \item We design a dedicated \textbf{FEALLM} for FEA, which extracts local facial features and combines them with low-level features from the visual encoder, enabling the MLLM to capture more detailed facial information.
    \item Extensive experiments demonstrate FEALLM's strong emotional analysis performance on FEABench and its impressive generalization capability, as evidenced by zero-shot evaluations on various datasets, including RAF-DB, AffectNet, BP4D, and DISFA.
\end{itemize}

\section{Related Work}

\subsection{Multimodal Large Language Models}

In the past two years, Multimodal Large Language Models (MLLMs) have rapidly advanced~\cite{yin2023survey}. MLLMs are designed to tackle a broad spectrum of multimodal tasks by integrating diverse inputs, such as visual data, into large language models using techniques like CLIP~\cite{radford2021learning}, and have proven to be highly versatile in understanding and processing complex multimodal information, with applications in tasks such as image captioning~\cite{bianco2023improving}, visual question answering~\cite{hu2024bliva,sterner2024few}, and other language-related capabilities~\cite{wu2024q}. By combining both visual and textual inputs, MLLMs can generate more accurate and contextually aware responses. Representative MLLMs include LLaVA~\cite{liu2024improved}, Qwen-VL~\cite{bai2023qwen}, mPLUG-Owl~\cite{ye2024mplug2}, InternVL~\cite{chen2024far}, and others. However, despite their success in various tasks, these MLLMs still face challenges in emotion understanding due to the lack of specialized training corpora, especially in tasks related to FEA. In this context, equipping MLLMs with ability to perceive and understand emotions is a highly valuable research direction.

\subsection{Multimodal Large Language Models for FEA}
So far, several pioneering studies have explored the application of MLLMs in FEA.
Lian \textit{et al.}~\cite{lian2024gpt} benchmarked common emotion datasets using GPT-4V.
Xing \textit{et al.}~\cite{xing2024emo} used Gemini to generate instruction data for FER, developing a facial-prior-informed MLLM.
Lan \textit{et al.}~\cite{lan2024expllm} incorporated chain-of-thought (CoT) prompts for FER to boost MLLMs' reasoning abilities.
Yang \textit{et al.}~\cite{yang2024emollm} focused on modeling and responding to complex human emotions using video-text data and specialized MLLM models.
Li \textit{et al.}~\cite{li2025facial} applied MLLM with tailored instructions and efficient tuning for facial affective behavior analysis.

Despite these preliminary explorations, there is still a lack of an MLLM that can deeply and comprehensively perform FEA. In contrast, we bridge this gap by constructing an FEA Instruction Dataset that includes not only accurate and aligned basic FE and AU descriptions but also complex emotional reasoning instructions, and by designing a specialized model, FEALLM, which effectively captures detailed facial information, thereby achieving strong emotional perception and reasoning abilities with robust generalization, contributing to a key step in applying MLLMs to FEA.

\begin{figure}[t]
    \centering
    \includegraphics[width=\linewidth]{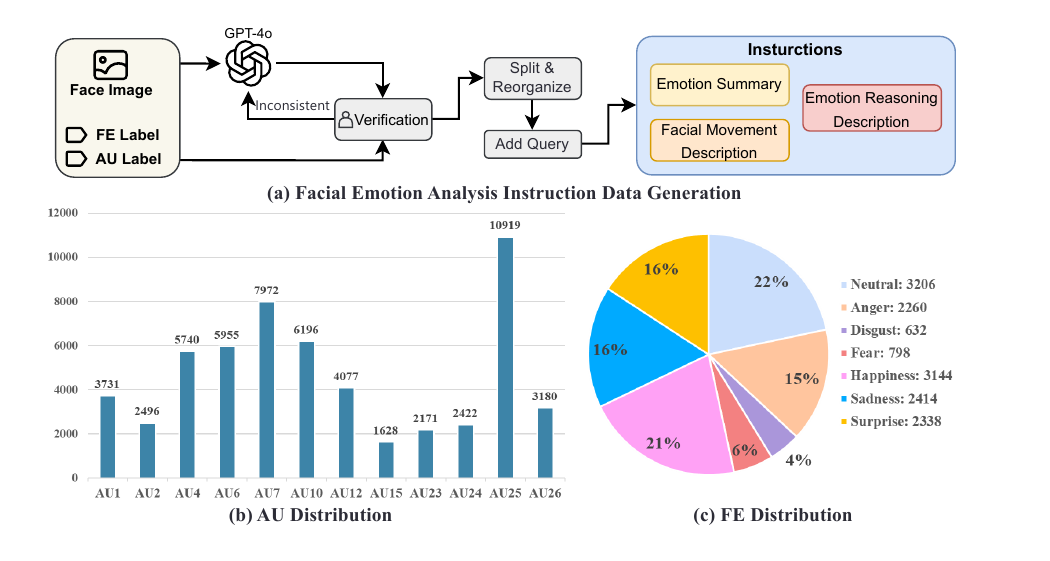}
    \caption{(a) The pipeline for generating FEA instruction data, and the distribution of (b) AUs and (c) FEs in the FEA dataset.}
    \label{fig2}
\end{figure}

\section{Data Construction}

In this section, we introduce the FEA Instruction Dataset and the brand-new benchmark FEABench. The former mainly consists of three types of instructions: emotion summaries, facial movement descriptions, and emotion reasoning descriptions, while the latter includes two primary evaluation tasks: FER and AUD. Our FEA Instruction Dataset is the first to provide accurate and aligned FE and AU descriptions and establish causal reasoning relationships between these two different granular levels of emotional representations, thereby enhancing the facial emotion perception and reasoning capabilities of MLLMs to promote their application in FEA.

\subsection{Data Annotation and Source}

Previous facial emotion instruction datasets typically relied on single emotion annotations (FE or AU), with missing labels generated using existing toolkits or GPT~\cite{li2025facial,lan2024expllm}. However, due to the inherent weaknesses of these tools in FEA, this approach often leads to unreliable labels, which negatively impact the overall quality of the dataset. Specifically, EmoLA~\cite{li2025facial} conducted a manual sampling review of GPT-generated missing labels and found that the accuracy of FE annotations was about 91\%, while the average F1 score for AU annotations was 76\%, with AU10 reaching only 58.1\%. Such potential noise in the annotations introduces inappropriate biases, reduces the reliability of the dataset, and consequently hinders the model's ability to understand facial emotions.

To overcome the aforementioned limitation, our FEA dataset is built upon the Aff-Wild2 dataset~\cite{kollias1811aff}, a large-scale in-the-wild dataset with both FE and AU annotations, including seven basic FE categories (Neutral, Anger, Disgust, Fear, Happiness, Sadness, Surprise) and 12 AUs (1, 2, 4, 6, 7, 10, 12, 15, 23, 24, 25, 26), covering common emotional states encountered in daily life. The FE and AU labels in the dataset have been annotated and validated by multiple domain experts, ensuring the reliability of the annotations.

We sample 16,227 facial images from the Aff-Wild2 dataset, all of which include FE and AU annotations. Of these, 14,892 images are used to construct the instruction dataset for training MLLMs, and 1,335 images are used for evaluation. The distribution of AUs and FEs in the constructed FEA dataset is shown in Fig.~\ref{fig2} (b) and (c). During the sampling process, we ensure that the training and evaluation data come from different subjects to properly assess the model's generalization ability.

\subsection{Instruction Data Generation}

Our instruction data not only includes aligned basic FE descriptions and AU descriptions but also incorporates emotional reasoning instructions that link AUs to FEs. By establishing causal relationships between specific, observable facial movements and abstract emotional states, these instructions clearly articulate the reasoning process from AU to FE, making it possible for the model to achieve a deeper understanding and perception of facial emotions.

The process of constructing instruction data is shown in Fig. \ref{fig2} (a). To streamline the generation of instruction data, we use GPT-4o~\cite{achiam2023gpt} to assist in creating structured textual descriptions for each facial image, where the facial image along with its FE and AU labels are provided as input. The prompt for instructing GPT-4o is as follows:
\begin{table}[h]\centering
\begin{minipage}{0.99\columnwidth}    \centering
\vspace{-1.7em}
\begin{tcolorbox} 
    \small
    \vspace{-0.3em}
    \PredSty{\texttt{<Image>}} The facial image expresses the emotion of \PredSty{\texttt{<fe\_label>}}, and the following Action Units (AUs) are activated: \PredSty{\texttt{<au\_label>}}. Please directly state the emotional label of the image with only one word, and then briefly describe the facial expression of the person in the image in one sentence to help understand the emotion. And then describe the character's facial movements based on the image and the activation of the AUs. Finally, explain how to derive the character's emotions from the AUs.  \\ 
     \vspace{-1.5em}
\end{tcolorbox}  
\vspace{-1.5em}
\end{minipage}
\end{table}


\noindent Notably, after obtaining the generated structured text, we validate it to ensure consistency with the original annotations of the corresponding facial image.

\begin{figure}[t]
    \centering
    \includegraphics[width=\linewidth]{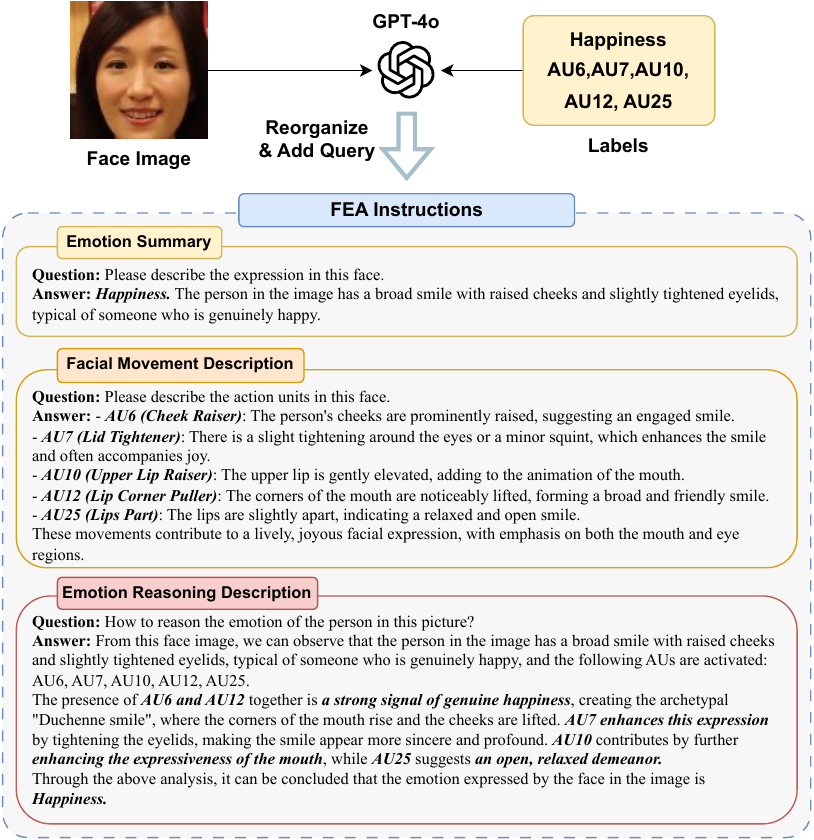}
    \caption{
    An example of the FEA Instruction Dataset.
    The face image and its corresponding labels are provided as input to GPT-4o to generate a structured description, which is then split and reorganized into three types of instructions: emotion summary, facial movement description, and emotion reasoning description.
    }
    \vspace{-1.5em}
    \label{fig3}
\end{figure}

The generated structured textual descriptions are divided into three parts: emotion summary, facial movement description, and emotion reasoning description. The emotion summary directly classifies and summarizes the individual's FE, while the facial movement description explains the facial activities triggered by the activated AUs. The emotion reasoning description clarifies how each activated AU influences the overall emotional expression. For each of the above types of textual descriptions, we carefully equip them with more than ten question templates to form the instruction data, with details provided in the Appendix. Additionally, we reorganize the emotion reasoning descriptions to enhance their logical coherence. An example of the organized instruction data is shown in Fig.~\ref{fig3}. These instruction data guide MLLMs in learning the basic FE and AU representations, as well as the logical reasoning process from AUs to FEs, thereby improving the model's overall performance in FEA.

\subsection{FEABench}

To promote synergistic development across FER and AUD tasks, based on the proposed FEA dataset, we introduce a brand-new benchmark FEABench. Unlike traditional benchmarks that typically train and evaluate these tasks separately, our FEABench simultaneously evaluates the model's ability to analyze both FE and AU, assessing its capacity to effectively uncover the intrinsic relationships between them.

The evaluation is conducted using the evaluation set from the FEA dataset. For the FER task, the goal is to classify the emotion presented on the face into one of the seven basic FE categories. Following \cite{EAC,VTFF}, Accuracy is used to measure the model's ability to recognize FE. The prompt for this task is randomly sampled from the question templates corresponding to the emotion summary, such as \textit{"Please describe the expression in this face."}


For the AUD task, the goal is to determine which AUs are activated on the face. The AUs predicted by the model are extracted from the generated text using regular expressions, and the model's ability to recognize AUs is measured using F1 scores~\cite{F1score}, following \cite{yuan2025auformer,FAN-Trans}. The prompt for this task is randomly sampled from the question templates corresponding to the facial movement description, such as \textit{"Please describe the action units in this face."}


\begin{figure*}[t]
    \centering
    \includegraphics[width=\textwidth]{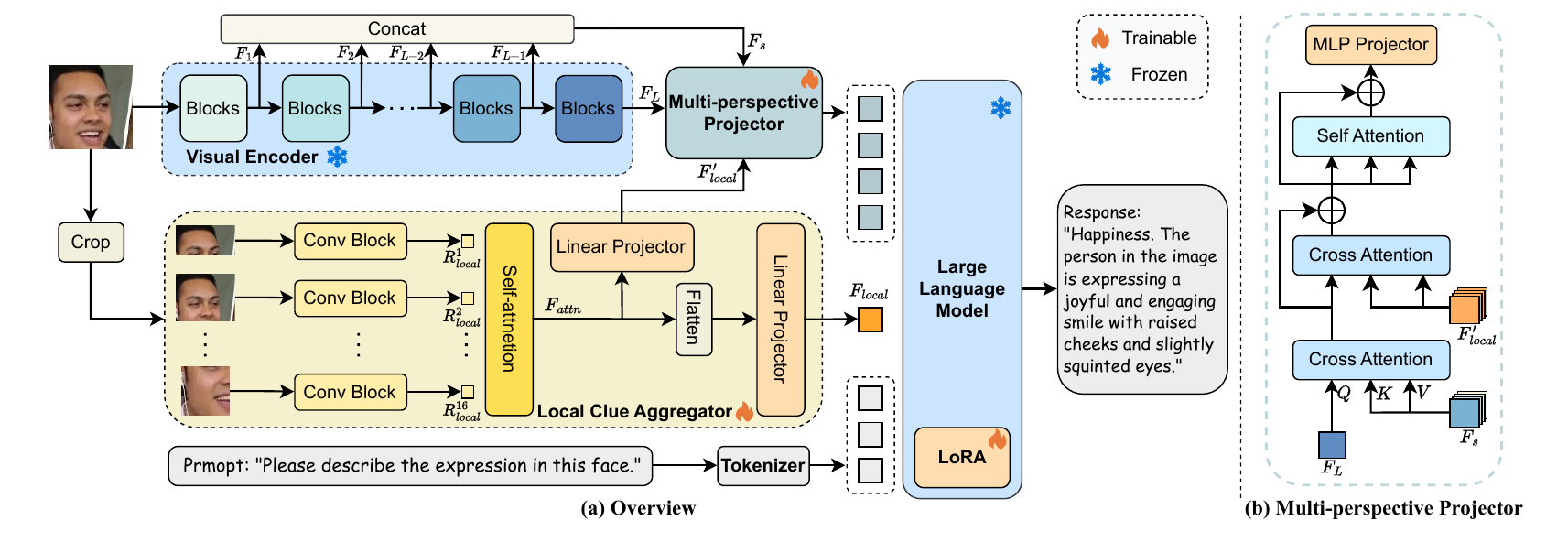}
    \caption{
    (a) The overview of the FEALLM architecture.
    (b) The Multi-perspective Projector integrates shallow features from the visual encoder and local region features into deep features to enhance the perception of low-level facial details.
    }
    \label{fig4}
\end{figure*}

\section{FEALLM}

In this section, we provide a detailed explanation of FEALLM, a model specifically designed for FEA tasks that can capture more detailed facial information. We will first describe its overall architecture, followed by an in-depth exploration of its key components.

\subsection{Overview}

The overview of FEALLM is shown in Fig.~\ref{fig4}. The proposed FEALLM architecture includes a visual encoder, a Local Clue Aggregator (LCA), a Multi-perspective Projector (MPP), and a Large Language Model (LLM). Each facial image is first input into both the visual encoder and the LCA for encoding. Then, these encoded representations are mapped into the token embedding space of the LLM via the MPP. The resulting tokens are concatenated with the provided instruction tokens and fed into the LLM for autoregressive generation, producing the corresponding response. Specifically, we use the pre-trained ViT-L/14 model from CLIP~\cite{radford2021learning} as the visual encoder. For the LLM, we adopt the pre-trained Vicuna-7B model~\cite{vicuna2023} from LLaVA-v1.5~\cite{liu2024improved}. Given the high training cost of LLMs, full fine-tuning of the entire model is impractical under resource constraints. Therefore, we fine-tune only the LCA, MPP, and the additional Low-Rank Adaptation (LoRA)~\cite{hu2021lora} module introduced under the condition of freezing the LLM's parameters, which allows for efficient adaptation while minimizing memory and computational overhead. 
Next, we provide a detailed description of the LCA and MPP modules.

\subsection{Local Clue Aggregator}

Local facial details are crucial for performing FEA, which general MLLMs typically struggle to capture. Hence, the Local Clue Aggregator (LCA) is introduced. However, facial images in the wild are often complex and diverse, making it extremely challenging to directly locate specific facial regions. 
To address this issue, we divide the image into multiple local regions by cropping parts from the original image in eight directions: the top, bottom, left, right, and the four corners. For each direction, we crop the image at half and three-quarters of the side length.
A visual illustration of the divide is provided in Appendix. This process generates a total of 16 local facial region images, which are then resized to $48\times48$ pixels.

Given these local regions, we first use a convolutional block containing four consecutive convolutional layers to extract features.
Let the local regions of the facial image be denoted as $I^{i}_{local} \in \mathbb{R}^{48\times 48\times 3}$, where $i \in \{1,2,...,16\}$.
The process of extracting local features can be presented as:
\begin{equation}
  R^{i}_{local} = \text{AvgPool}(\text{ConvBlock}(I^{i}_{local})),
\end{equation}
where $R^{i}_{local}\in\mathbb{R}^{d}$, $d$ is the number of channels in the convolutional kernel, and $\text{AvgPool}$ represents average pooling, which is used to compress the spatial dimensions of the features. Then, the local features, represented as $\{R_{local}^{1}, R_{local}^{2}, ..., R_{local}^{16}\}$, are stacked together, resulting in $R_{local} \in \mathbb{R}^{16\times d}$.

Since different local regions contribute differently to facial emotion analysis, we use the self-attention mechanism to adaptively learn and assign weights to these regions, which can be formulated as:
\begin{equation}
     F_{attn} = \text{Attn}(Q,K,V)=\text{Softmax}(\frac{QK^T}{\sqrt{d}})V,
\end{equation}
where $Q$, $K$, and $V$ are all $R_{local}$.

Finally, the refined local features $F_{attn}$ are flattened and aggregated through a linear layer, mapping them into the token embedding space of LLM, which can be formulated as:
\begin{equation}
   F_{local}=\text{Linear}(\text{Flatten}(F_{attn})),
\end{equation}
where $F_{local} \in \mathbb{R}^{d^\prime}$ will be concatenated with the instruction tokens to be input into the LLM, and $d^\prime$ is the dimension of the token embedding space. The illustration of the LCA module is shown in the yellow area of Fig.~\ref{fig4} (a).

\subsection{Multi-perspective Projector}

Some studies~\cite{cao2024mmfuser,yao2024dense} have shown that deep features from visual encoders are capable of capturing high-level semantic information but may neglect low-level details such as edges and textures, which are typically captured by shallow features and play a critical role in effective facial emotion perception. Clearly, fully leveraging shallow features helps the model better capture facial details, enhancing its ability to analyze facial emotions.

Therefore, the Multi-perspective Projector (MPP) is introduced, as illustrated in Fig.~\ref{fig4} (b). Specifically, the MPP first extracts $L$ feature maps from different layers of the visual encoder, denoted as $F_l$, where $l = \{1, 2, ..., L\}$. Then, the deep feature $F_L$ is used as the Query, and the shallow features $F_s = \text{Concat}(F_1, F_2, ..., F_{L-1})$ are used as the Key and Value, enabling the dynamic extraction of missing low-level details from the shallow features through cross-attention mechanism, as formulated below:
\begin{equation}
    F_{shallow}^{fuse} = \text{CrossAttn}(F_L, F_s, F_s).
\end{equation}

After enriching $F_L$ with shallow features, we further fuse the obtained $F_{shallow}^{fuse}$ with the local features extracted from the LCA module to perceive the details of the local regions. The local features $F_{attn}$ are first passed through a linear projector, as formulated below:
\begin{equation}
    F^\prime_{local} = \text{Linear}(F_{attn}).
\end{equation}
Next, we perform cross-attention with dynamic residual connections between the shallow features-enhanced $F_{shallow}^{fuse}$ and the projected local features $ F^\prime_{local}$, as formulated below:
\begin{equation}
    F_{local}^{fuse} = \text{CrossAttn}(F_{shallow}^{fuse}, F^\prime_{local}, F^\prime_{local}) + \gamma_1 \cdot F_{shallow}^{fuse},
\end{equation}
where $\gamma_1$ is a learnable scaling factor.

To further improve feature interactions, we also apply a round of self-attention with dynamic residual connections to the obtained $F_{local}^{fuse}$, as formulated below:
\begin{equation}
    F_{fuse} = \text{SelfAttn}(F_{local}^{fuse}) + \gamma_2 \cdot F_{local}^{fuse}
\end{equation}
where $\gamma_2$ is also a learnable scaling factor.

Finally, the enriched and fused multi-perspective features are passed through a two-layer MLP projector, which aligns them into the LLM's text embedding space:
\begin{equation}
    F_{vision} = \text{MLP}(F_{fuse}).
\end{equation}

\begin{table*}[t]
\centering

\caption{Comparison of facial emotion analysis results on the proposed FEABench. Accuracy and F1 score are used as metrics for the FER and AUD tasks respectively. All scores are in \%. The best and second best results for each column are \textbf{bolded} and \underline{underlined}, respectively.}

\resizebox{\linewidth}{!}{
\begin{tabular}{lc|c|cccccccccccc|c}
\toprule
\multirow{2}{*}{Methods} &  & \multirow{2}{*}{FE} & \multicolumn{13}{c}{AU} \\
\cmidrule(lr){4-16}
    & & ~ & {1} & {2} & {4} & {6} & {7} & {10} & {12} & {15} & {23} & {24} & {25} & {26} & {\textbf{Avg.}}\\
\midrule

Qwen-VL-Chat~\cite{bai2023qwen}    &\textit{zero-shot}      & {45.71} & {9.40} & {1.10} & {4.90} & {0.00} & {0.21} & {2.20} & {1.22} & {0.00} & {0.00} & {0.00} & {0.00} & {0.00} & {1.59}\\
mPLUG-Owl2-7B~\cite{ye2024mplug2} &\textit{zero-shot} & {45.18} & {2.35} & {3.19} & {1.14} & {0.37} & {1.22} & {0.50} & {0.82} & {0.00} & {0.00} & {0.00} & {0.19} & {0.96} & {0.89}\\
internVL2-8B~\cite{chen2024far} &\textit{zero-shot} & {42.87} & {0.00} & {0.00} & {0.00} & {47.94} & {6.69} & {53.44} & {51.29} & {0.00} & {0.00} & {1.29} & {42.99} & {17.40} & {18.48}\\
MiniCPM-o-2.6-8B~\cite{hu2024minicpm} &\textit{zero-shot} & {57.28} & {2.33} & {1.95} & {29.58} & {9.67} & {13.28} & {9.03} & {1.21} & {2.41} & {0.00} & {0.00} & {0.19} & {0.00} & {5.80}\\
LLaVA-1.5-7B~\cite{liu2024improved}   &\textit{zero-shot}      & {52.65} & {28.69} & {2.08} & {0.37} & {3.30} & {0.00} & {0.00} & {0.00} & {0.00} & {0.00} & {0.00} & {0.00} & {0.00} & {2.87}\\
LLaVA-1.5-7B (w/ LoRA)~\cite{liu2024improved}    &\textit{fine-tuning}      & \underline{59.97} & \underline{37.25} & \textbf{33.55} & \underline{83.01} & \underline{76.15} & \textbf{78.09} & \underline{74.00} & \underline{78.69} & \underline{24.16} & \textbf{12.31} & \underline{53.40} & \underline{86.72} & \underline{32.21} & \underline{55.79}\\
\midrule
\rowcolor{gray!30}
\textbf{FEALLM (ours)}   &\textit{fine-tuning}    & \textbf{67.36} & \textbf{53.51} & \underline{33.33} & \textbf{87.99} & \textbf{77.92} & \underline{76.94} & \textbf{78.56} & \textbf{80.68} & \textbf{25.50} & \underline{6.72} & \textbf{66.67} & \textbf{88.18} & \textbf{39.17} & \textbf{59.60}\\
\bottomrule
\end{tabular}
}

\label{tab:FEABench}
\end{table*}
\begin{table}[t]
\centering
\caption{Zero-shot evaluation on the FER datasets RAF-DB and AffectNet. Accuracy (in \%) is used as the metric. The best and second best results for each column are \textbf{bolded} and \underline{underlined}, respectively.}
\resizebox{0.8\linewidth}{!}{
\begin{tabular}{l|c|c}
\toprule
{Methods} & {RAF-DB} & {AffectNet} \\
\midrule

Qwen-VL-Chat~\cite{bai2023qwen}   & {57.53}  & {39.94}\\
mPLUG-Owl2-7B~\cite{ye2024mplug2}  & {44.20} & {37.51}\\
internVL2-8B~\cite{chen2024far}  & {50.62} & \textbf{42.04}\\
MiniCPM-o-2.6-8B~\cite{hu2024minicpm} & {59.58} & {40.28} \\
LLaVA-1.5-7B~\cite{liu2024improved}    & {55.25} & {39.66}\\
LLaVA-1.5-7B (w/ LoRA)~\cite{liu2024improved}   & {\underline{63.40}} & {36.03}\\
\midrule
\rowcolor{gray!30}
\textbf{FEALLM (ours)}    & \textbf{69.95} & {\underline{41.91}}\\
\bottomrule
\end{tabular}
}

\label{tab:rafdb_affectnet}
\end{table}

\section{Experiments}

\subsection{Implementation Details}
We initialize all the frozen weights of FEALLM with LLaVA-v1.5-7B~\cite{liu2024improved}. Specifically, we adopt CLIP-ViT-L/14-336px~\cite{radford2021learning} as the visual encoder and Vicuna-7B~\cite{vicuna2023} as the LLM. For the convolutional layers in the LCA module, we set the kernel size to $3\times 3$ and the number of channels $d$ to 64. For the MPP module, we uniformly sample $L = 5$ feature maps from the visual encoder. Specifically, we extract features from layers [3, 8, 13, 18] as shallow features, and the features from layer 23 as deep features. The attention mechanism in MPP module is implemented using deformable attention~\cite{zhu2020deformable}.

We adopt a two-stage training strategy. In the pre-training stage, we freeze the LLM and vision encoder, and train the LCA and MPP modules on the LLaVA-LCS-558K~\cite{liu2024improved} dataset to align the visual features with the LLM's embedding space. In the fine-tuning stage, we incorporate LoRA modules into the LLM, setting the LoRA rank to 128, and fine-tune the LCA module, MPP module, and LoRA on our FEA dataset. During the pre-training stage, the learning rate is set to $1e^{-3}$ with a batch size of 64. During the fine-tuning stage, the learning rates for the LCA and MPP modules are set to $2e^{-5}$, while the learning rate for LoRA is set to $2e^{-4}$, with a batch size of 16. Both stages are trained for one epoch. All experiments are performed using two NVIDIA A100 Tensor Core GPUs.

\subsection{Comparison on FEABench}

We first conduct comparative experiments on the proposed FEABench with popular MLLMs, and the results are shown in Tab.~\ref{tab:FEABench}. It can be observed that, although existing general-purpose MLLMs demonstrate excellent performance across multiple visual tasks, their performance in FEA tasks remains unsatisfactory, especially in the AUD task, where the F1 score for some AUs is even 0. It is worth noting that, in addition to the MLLMs listed in Tab.~\ref{tab:FEABench}, we also evaluate MiniGPT4-v2~\cite{chen2023minigpt} and InstructBLIP~\cite{dai2023instructblip}, but these MLLMs fail to provide meaningful responses when faced with FEA tasks, essentially crashing. The issues described above mainly arise from the lack of FEA-related instruction data in the current training corpora, making the construction of an FEA instruction dataset essential to facilitate the application of MLLMs in FEA. Evidently, as shown in the penultimate row of Tab.~\ref{tab:FEABench}, the LLaVA-1.5 model with LoRA shows a significant leap in its ability to handle FEA tasks after fine-tuning on our FEA dataset, demonstrating the effectiveness of the proposed FEA dataset. However, it still lags behind our FEALLM in both the FER and AUD tasks, which can be attributed to the LCA and MPP modules introduced in FEALLM. These modules facilitate the effective utilization of local facial features and low-level features from the visual encoder, enabling FEALLM to better capture facial details and significantly improve facial emotion perception.

\subsection{Zero-shot Evaluation}

In addition to the proposed FEABench, we also conduct zero-shot evaluations on two FER datasets: RAF-DB~\cite{RAF-DB} and AffectNet~\cite{AffectNet}, as well as two AU detection datasets: BP4D~\cite{BP4D} and DISFA~\cite{DISFA}, to further explore the generalization capability of MLLMs across multiple datasets and a wide range of scenarios. A detailed introduction to all the datasets involved can be found in the Appendix.

For the FER task, we directly use the provided test sets from the RAF-DB and AffectNet datasets for evaluation and measure the model's performance using Accuracy, with the results shown in Tab.~\ref{tab:rafdb_affectnet}. It can be observed that, on the RAF-DB dataset, our FEALLM significantly outperforms other MLLMs by at least 6.55\%, achieving the highest accuracy of 69.95\%, demonstrating its superior performance in recognizing FE. Moreover, on the more challenging AffectNet dataset, the significant potential domain shift between it and our training data leads to a situation where, although FEA instruction tuning enhances the model’s ability to perceive and understand facial emotions, it may still fail to fully align with the broader complexity of AffectNet, as evidenced by the performance degradation of the LLaVA-1.5 model fine-tuned with LoRA. Nevertheless, FEALLM still maintains a leading position, indicating that our model’s ability to capture fine-grained facial details helps mitigate the domain shift, demonstrates robust generalization in handling diverse environments, and preserves its competitive advantage.


\begin{table}[t]
    \centering
    \caption{Ablation study on the three types of instruction data.}
    \resizebox{\linewidth}{!}{
        \begin{tabular}{ccc|cc}
            \toprule
            Emotion Summary & Facial Movement & Emotion Reasoning & {FE} & {AU}\\
            \midrule
            \checkmark & - & - & 59.37 & 0.00\\
            - & \checkmark & - & 34.58 & 58.70\\
            \checkmark & \checkmark & - & 63.65 & 58.79\\
            \rowcolor{gray!30}
            \checkmark & \checkmark & \checkmark & \textbf{67.36} & \textbf{59.60}\\
            \bottomrule
        \end{tabular}
    }
    
    \label{tab:ablation_data}
\end{table}
    
\begin{table}[t]
    \centering
    \caption{Ablation study on various modules.}
    \resizebox{\linewidth}{!}{
        \begin{tabular}{cc|cc}
            \toprule
            Local Clue Aggregator & Multi-perspective Projector & {FE} & {AU}\\
            \midrule
            - & - & 59.97 & 55.79 \\
            \checkmark & - & 61.69 & 57.10\\
            - & \checkmark & 65.57 & 58.43\\
            \rowcolor{gray!30}
            \checkmark & \checkmark  & \textbf{67.36} & \textbf{59.60}\\
            \bottomrule
        \end{tabular}
    }
    
    \label{tab:ablation_model}
\end{table}
\begin{table*}[t]
\centering
\caption{Zero-shot evaluation on the AUD datasets BP4D and DISFA. F1 score (in \%) is used as the metric. The best and second best results for each column are \textbf{bolded} and \underline{underlined}, respectively.}
\resizebox{\linewidth}{!}{
\begin{tabular}{l|cccccccccc|c|ccccccc|c}
\toprule
\multirow{2}{*}{Methods}  & \multicolumn{11}{c|}{BP4D} & \multicolumn{8}{c}{DISFA}  \\
\cmidrule(lr){2-12}
\cmidrule(lr){13-20}
 & {AU1} & {AU2} & {AU4} & {AU6} & {AU7} & {AU10} & {AU12} & {AU15} & {AU23} & {AU24} & {\textbf{Avg.}} & {AU1} & {AU2} & {AU4} & {AU6} & {AU12} & {AU25} & {AU26} & {\textbf{Avg.}} \\
\midrule

Qwen-VL-Chat~\cite{bai2023qwen}    & {29.27} & {17.77} & {22.53} & {6.28} & {2.31} & {17.32} & {6.53} & {2.61} & {1.18} & {1.68} & {10.75} & {5.80} & {3.73} & {12.01} & {0.90} & {1.61} & {0.86} & {0.84} & {3.68} \\
mPLUG-Owl2-7B~\cite{ye2024mplug2} & {12.61} & {7.26} & {6.01} & {7.12} & {6.00} & {5.60} & {5.25} & {6.63} & {2.61} & {2.99} & {6.21} & {6.03} & {0.00} & {0.49} & {0.92} & {1.15} & {0.85} & {0.00} & {1.35} \\
internVL2-8B~\cite{chen2024far} & {12.93} & {5.15} & {2.36} & {63.37} & {8.34} & {22.53} & {71.49} & {0.00} & {0.00} & {0.00} & {18.62} & {5.65} & {0.00} & {0.00} & {27.30} & {27.75} & {43.34} & {16.29} & {17.19} \\
MiniCPM-o-2.6-8B~\cite{hu2024minicpm} & \underline{30.23} & {8.49} & {29.94} & {55.20} & {41.76} & {29.34} & {6.84} & {0.74} & {0.00} & {0.00} & {20.25} & {9.03} & {0.94} & {21.03} & {16.03} & {4.76} & {0.29} & {0.00} & {7.44} \\
LLaVA-1.5-7B~\cite{liu2024improved}   & \textbf{34.61} & {5.39} & {2.12} & {2.17} & {0.86} & {1.26} & {1.46} & {0.39} & {0.00} & {1.28} & {4.66} & {5.68} & {0.00} & {1.02} & {0.00} & {0.00} & {0.00} & {0.00} & {0.96}\\
LLaVA-1.5-7B (w/ LoRA)~\cite{liu2024improved}   & {23.98} & \textbf{24.82} & \underline{44.10} & \textbf{74.83} & \underline{74.96} & \textbf{80.81} & \textbf{85.58} & \underline{19.40} & \underline{18.82} & \underline{7.81} & \underline{45.51} & \underline{30.05} & \underline{25.21} & \underline{60.54} & \underline{28.52} & \textbf{60.20} & \textbf{52.34} & \underline{23.52} & \underline{40.06} \\
\midrule
\rowcolor{gray!30}
\textbf{FEALLM (ours)}   & {25.73} & \underline{21.98} & \textbf{46.78} & \underline{72.90} & \textbf{75.09} & \underline{79.28} & \underline{84.88} & \textbf{21.43} & \textbf{19.53} & \textbf{12.20} & \textbf{45.98} & \textbf{36.90} & \textbf{29.69} & \textbf{70.41} & \textbf{30.15} & \underline{54.50} & \underline{51.87} & \textbf{24.73} & \textbf{42.61} \\
\bottomrule
\end{tabular}}

\label{tab:bp4d&disfa}
\end{table*}


For the AUD task, considering that the BP4D and DISFA datasets are annotated based on continuous video frames, we perform 2\% uniform sampling on the entire dataset for more efficient testing. Notably, we only evaluate the AU types shared between the target datasets and our FEA Instruction Dataset. The model’s performance is measured using the F1 score, with the results shown in Tab.~\ref{tab:bp4d&disfa}. 
As seen, our FEALLM model outperforms other MLLMs across both the DISFA and BP4D datasets. While some models like LLaVA-1.5 fine-tuned with LoRA excel in specific AUs, FEALLM maintains strong performance across a broad range of AUs, achieving the highest average F1 score. These results underscore FEALLM’s robustness and exceptional ability to detect AUs effectively, even in zero-shot settings, highlighting its potential for advanced facial emotion analysis.

\begin{figure}[t]
    \centering
    \includegraphics[width=\linewidth]{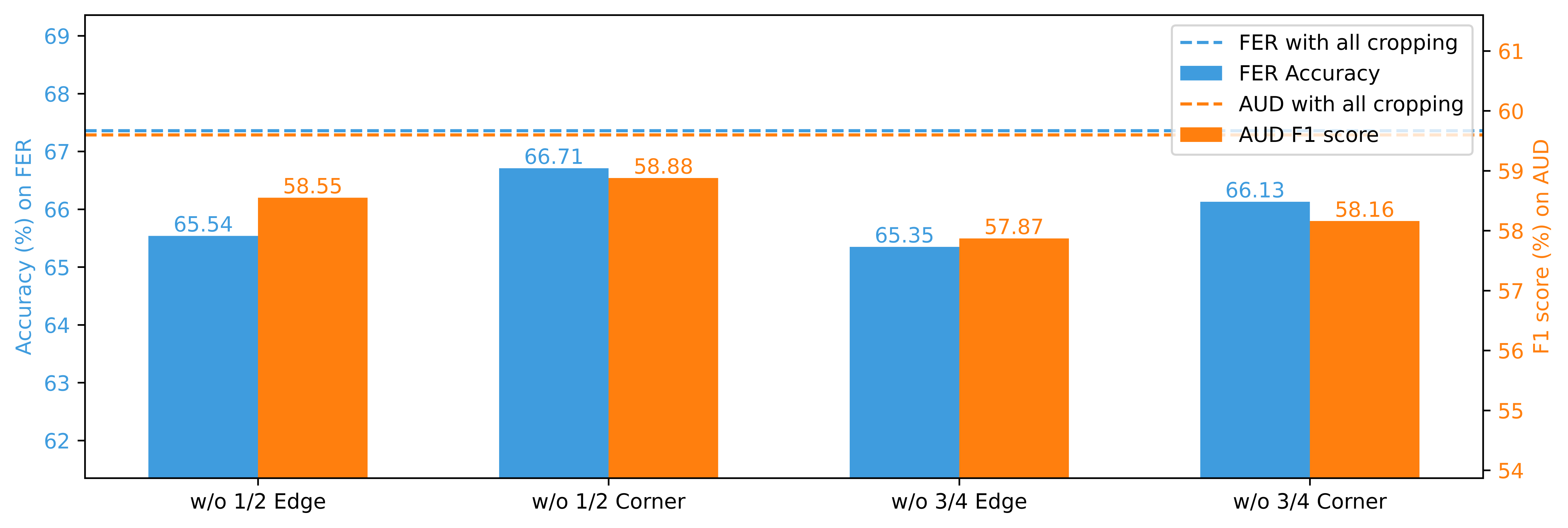}
    \caption{Ablation study on cropping types in the Local Clue Aggregator module.}
    \label{fig:FER_AUD_crop}
\end{figure}

\subsection{Ablation Study}
We conduct ablation studies on FEABench to explore the effects of instruction data types and key modules in FEALLM.

\textbf{Effectiveness of Different Instruction Data.}
The ablation study results in Tab.~\ref{tab:ablation_data} demonstrate the impact of different types of instruction data on FEA tasks. It can be observed that using only emotion summary instructions improves FER accuracy but fails to detect AUs. On the other hand, facial movement description instructions help with better AU detection but lead to a lower FER accuracy, highlighting the limitation of focusing solely on facial movements. It is gratifying to note that when both emotion summary and facial movement description instructions are incorporated together, the model's performance in both FER and AUD tasks improves, validating the complementary effect between these two tasks that recognize different levels of emotional representations, which aligns with psychological consensus~\cite{ekman1978facial}. Similarly, \cite{li2025facial} also attempted to train these two tasks together but failed to achieve synergy, which ultimately stems from the misalignment of their FE and AU-related instruction data. Furthermore, the introduction of emotion reasoning instructions significantly enhances the model's ability to link facial movements with emotional states, leading to substantial progress in both FER and AUD tasks, which demonstrates that incorporating emotion reasoning instructions not only strengthens the model's understanding and perception of emotions but also enables it to more effectively integrate facial cues and uncover the intrinsic relationships between tasks, thereby improving overall FEA performance.

\begin{figure}[t]
    \centering
    \includegraphics[width=\linewidth]{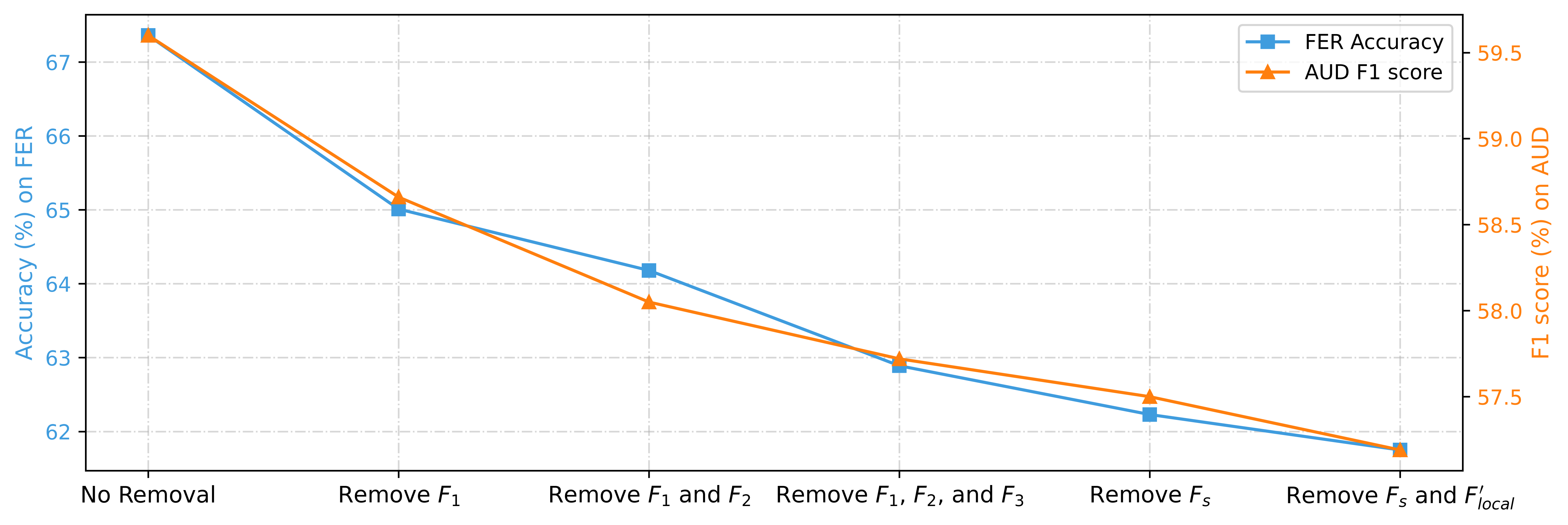}
    \caption{Ablation study on shallow and local features in the Multi-perspective Projector module.}
    \label{fig:Layer}
    \vspace{-1.0em}
\end{figure}

\begin{figure*}[t]
    \centering
    \includegraphics[width=0.95\linewidth]{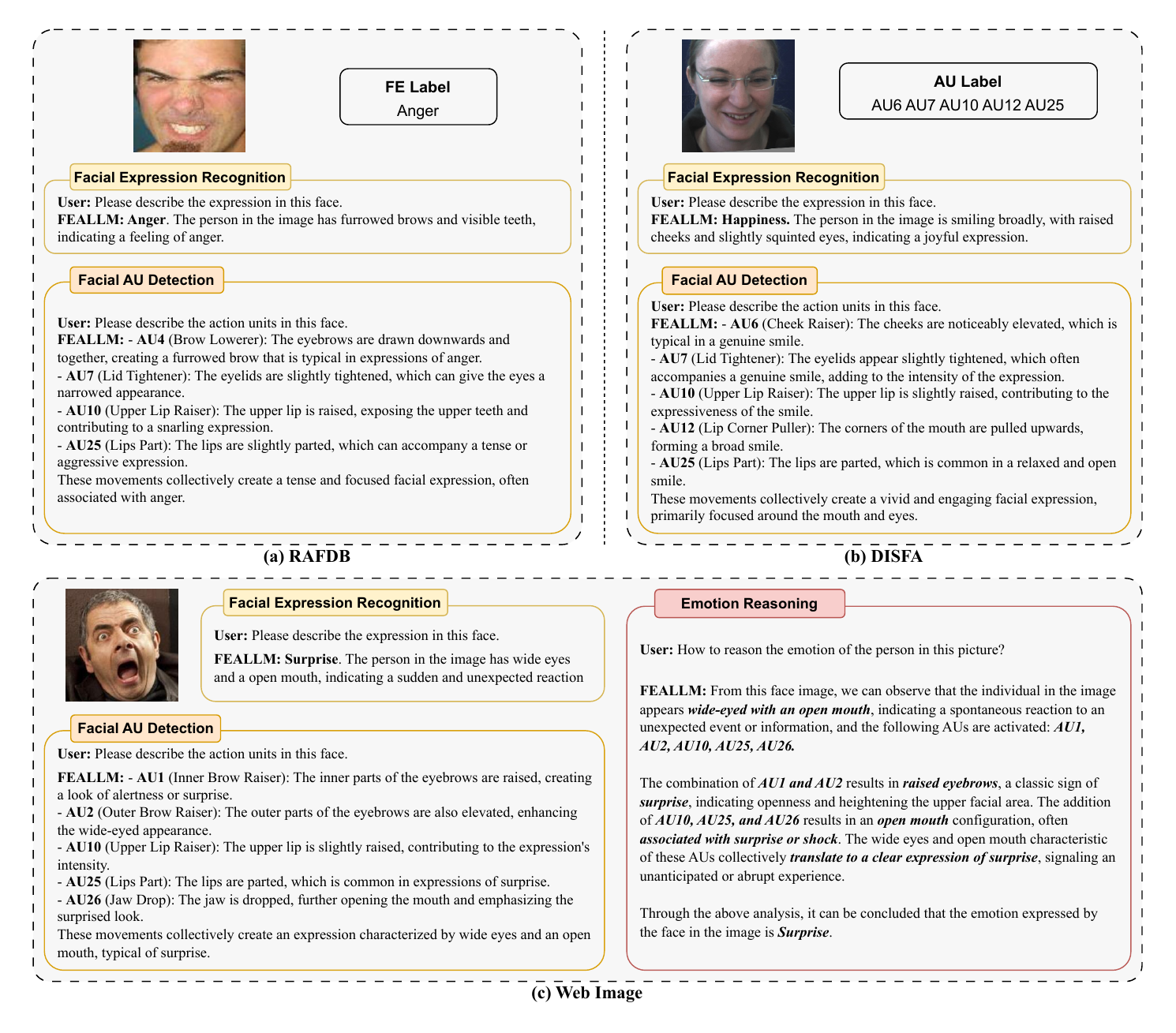}
    \caption{Visualization examples of FEALLM in FEA tasks. The face image (a) is from the FER dataset RAF-DB, (b) is from the AUD dataset DISFA, and (c) is from the web.}
    \label{vis}
\end{figure*}

\textbf{Effectiveness of Model Components.}
The ablation study results in Tab.~\ref{tab:ablation_model} highlight the contributions of the different modules introduced in FEALLM to FEA tasks. 
By incorporating the LCA module, the model becomes more capable of focusing on local areas of the face, which enhances its ability to detect subtle facial movement details.
Alternatively, adding the MPP module also boosts performance, indicating that shallow low-level features from visual encoder are critical for capturing essential facial cues related to FEA.
When both modules are used together, the model benefits from capturing detailed local features while leveraging shallow features for broader context, resulting in the best performance and demonstrating that integrating both modules enhances the model's FEA capability. 
To further investigate the effectiveness of the features involved in LCA and MPP, we ablate by removing certain features from the complete architecture. 
Specifically, Fig. \ref{fig:FER_AUD_crop} shows the results of removing one of the four cropping types (classified by direction and size) from the LCA, while Fig. \ref{fig:Layer} presents the results of progressively removing shallow features and local features used for interaction enhancement in the MPP. It can be observed that removing any involved features negatively impacts both FER and AUD tasks, highlighting their essential role and necessity in capturing key facial details and low-level texture information.

\subsection{Qualitative Results}

We present qualitative results to provide a deeper understanding of FEALLM's powerful capabilities in FEA. Specifically, for the FER dataset, given a facial image, our model not only accurately predicts the FE but, more importantly, captures detailed facial movements, detecting the activated AUs, as shown in Fig.~\ref{vis} (a). This highlights the model's ability to go beyond simple classification and also understand the underlying facial dynamics that contribute to emotional expression. Similarly, for the AUD dataset, our model excels not only in detecting AUs but also in precisely generating the corresponding FE labels, as shown in Fig.~\ref{vis} (b). These examples demonstrate the synergy between FER and AUD, indicating that our model can effectively integrate both tasks, leading to a more complete understanding and perception of facial emotional states. Excitingly, we further evaluate our model on facial images from the web, where it demonstrates not only the ability to correctly predict the FE and AUs of the given face but also to provide a reasonable reasoning process, as shown in Fig.~\ref{vis} (c). This showcases the strong robustness and generalization capability of our model, highlighting its ability to effectively handle diverse, real-world facial images. More qualitative results can be found in the Appendix.

\section{Conclusion}

In this paper, we introduce a novel approach to FEA that overcomes the limitations of traditional methods and existing MLLMs. By constructing the FEA Instruction Dataset and FEABench, we provide accurate FE and AU descriptions with causal reasoning instructions, enhancing the understanding of their relationships. Our model, FEALLM, captures fine-grained facial details, improving performance in FEA tasks. Extensive experiments on multiple datasets demonstrate its strong capability and generalization in both FER and AUD tasks.
Future work will focus on further expanding the FEA Instruction Dataset and exploring multimodal extensions, including incorporating video and audio, to further enhance the model's practicality in diverse, real-world scenarios.


\bibliographystyle{ACM-Reference-Format}
\bibliography{sample-base}

\end{document}